\documentclass[11pt,a4paper]{article}
\usepackage{fontspec}
\usepackage[a4paper,margin=1in]{geometry}
\usepackage{amsmath,array,booktabs,tabularx,graphicx}
\usepackage[unicode,hidelinks]{hyperref}
\hypersetup{pdftitle={Joint CTC/Attention Decoding and Repertoire Priors for Staff-Level Recognition of Medieval Plainchant},pdfauthor={Robert Klugseder}}
\date{}

\title{Joint CTC/Attention Decoding and Repertoire Priors for Staff-Level Recognition of Medieval Plainchant}

\author{\href{https://www.oeaw.ac.at/de/acdh/institut/team/mitarbeitende/robert-klugseder}{Robert Klugseder}\\{\normalsize Austrian Academy of Sciences (ACDH), Vienna}\\{\normalsize\href{mailto:robert.klugseder@oeaw.ac.at}{robert.klugseder@oeaw.ac.at}}}

\begin{document}

\maketitle

\begin{abstract}
We study a staff-level recogniser for medieval plainchant that transcribes staff positions and clefs. It combines a convolutional recurrent neural network (CRNN) encoder, a connectionist temporal classification (CTC) head and a Transformer decoder that can be initialised from a melody model trained on the melodies of the Cantus Database; at reading time, completed hypotheses can additionally be rescored with the melody model and with a score against other manuscripts' versions of the same chant. On a fixed, folio-disjoint test set of 3,084 staves, the CTC-only model reaches 7.10 \% token error (mean of three seeds); the hybrid with a neutral decoder and joint rescoring reaches 5.84 \%, and melody initialisation of the decoder lowers this to 5.65 \%. On known manuscripts the repertoire priors make no measurable difference. In leave-one-source-out folds for three early manuscripts, both priors together lower the melody error by 1.10 to 3.67 percentage points in each of three seeds, mostly by changing the initial clef. In the examined samples of changed clefs, image judgments confirm 41 corrections against 2 errors in the two earliest held-out manuscripts, while on known manuscripts the priors caused more clef errors than they corrected (9 against 5). We therefore propose source familiarity as a deployment heuristic for clef changes. On the same staves and with the same metric, we did not detect a difference from a fine-tuned vision-language model of about one billion parameters on known manuscripts; on five unseen manuscripts the vision-language model is ahead (1.31 \% against a CRNN seed mean of 1.82 \% pitch-position error; interval for seed 0 $-$1.03 to +0.06 points).
\end{abstract}

\section{Introduction}
Optical music recognition (OMR) for medieval chant manuscripts has to cope with many notational families, scribal hands and states of preservation, while ground truth exists only for a small part of the surviving sources. Chant differs from most other repertoires in one respect that matters for recognition: a large part of every antiphoner or gradual belongs to a shared repertoire whose texts and melodies are documented across hundreds of sources in the Cantus Database \cite{cantusdb}, many of them with melodies encoded in Volpiano \cite{helsen2011}. A human reader uses this knowledge constantly. This article asks under which conditions a recogniser can use it too, and what it gains from it compared with what it gains from its architecture and its training data.

We work at the level of one staff and transcribe the position of every note on the staff (line or space, counted from the bottom) together with the clef as written, not pitches. Pitches follow from positions and clefs; keeping the two apart separates what is seen from its musical interpretation and makes clef errors visible as such.

Our contributions are empirical; the main architectural components are taken from speech and handwriting recognition and named as such:

\begin{enumerate}
\item An ablation over three seeds of a hybrid recogniser for chant: a CRNN encoder \cite{calvo2018,calvo2019} with a CTC head and a Transformer decoder, trained jointly and used for joint rescoring of attention-beam hypotheses \cite{kim2017,hori2017}, with the decoder initialised from a melody model trained on chant melodies. The ablation separates the effect of the hybrid with joint rescoring from that of melody pre-training.
\item A concordance score built from other manuscripts' versions of the same chant, evaluated together with rescoring by an external melody model (in the sense of shallow fusion \cite{gulcehre2015}, here applied to completed hypotheses) on known manuscripts, in six leave-one-source-out folds, with three seeds for three folds and one seed for each of the remaining folds, and by a blind check against the images. It shows where repertoire knowledge helps (clefs of unseen early manuscripts) and where it harms (clefs of manuscripts the model knows), and it suggests a heuristic for its use.
\item A comparison with LightOnOCR-3-1B \cite{lightonocr2026} fine-tuned for chant \cite{klugseder2026lighton}, on identical staves, with the same reference and metric and under the same rule for held-out manuscripts (the training data differ; Section 6.4), and results on the three real datasets of the AMNLT benchmark \cite{amnlt}.
\end{enumerate}

Code and models are described in the availability statement.

\section{Related work}
\textbf{Staff-level OMR.} Treating a staff image as a sequence transcription problem, with a convolutional recurrent network and CTC loss, goes back to Calvo-Zaragoza and Rizo for monophonic printed scores \cite{calvo2018} and was carried over to handwritten mensural notation \cite{calvo2019}. For square notation, Wick, Hartelt and Puppe detect staves, symbols and melodies with fully convolutional networks \cite{wick2019}. Kurek and Hajič convert the synthetic and the three real AMNLT corpora into one GABC-derived encoding, clean near-duplicates across the dataset splits and train shared Document Attention Network variants; they additionally fine-tune on the three real corpora, while the synthetic corpus is evaluated with the shared model \cite{kurek2026}. Their work and ours are complementary: they pool corpora into one image-to-sequence model, while we add repertoire knowledge at reading time; because of the different encodings, their error rates are not directly comparable with ours.

\textbf{Joint CTC/attention.} Training a CTC head and an attention decoder on a shared encoder with a weighted loss, and scoring decoder hypotheses with both at decoding time, was introduced for speech recognition \cite{kim2017,hori2017}; rescoring sequence-to-sequence text-line recognisers with CTC prefixes is its counterpart in handwriting recognition \cite{wick2022}. We adopt the training objective; at reading time we do not integrate CTC prefix scores into the search but rescore completed attention-beam hypotheses (Section 4.3), and measure what this adds for chant.

\textbf{Language models and background knowledge.} Shallow fusion adds the log-probability of an external language model to the decoder score \cite{gulcehre2015}. For square notation, Hartelt and Puppe add rule-based background knowledge (overlap of notes and text, clefs, ligatures, staff position) to a segmentation-based pipeline and report improved symbol recognition \cite{hartelt2022}; a later paper describes the complete pipeline for historic chants \cite{hartelt2024}. In a preliminary attempt reported in the same 2022 article, an n-gram model of pitches applied as post-correction introduced about as many errors as it corrected \cite[p.~15]{hartelt2022}. Their knowledge acts on symbols after recognition; ours rescores whole candidate readings with a learned melody model and with versions of the same chant, and is evaluated on held-out manuscripts.

\textbf{Aligned music and lyrics, and vision-language models.} AMNLT \cite{amnlt} defines the joint transcription of notes, lyrics and their alignment for chant and compares divide-and-conquer CRNNs, unfolding approaches and image-to-sequence models (DAN, SMT). The CRNN studied here, trained with CTC \cite{graves2006} and an attention decoder, reads notes only; we compare it on the AMNLT test sets by pitch-position error, and with a vision-language model that reads notes and lyrics, on the same staves.

\section{Data}
\subsection{Target encoding}
A target is a sequence of tokens: clefs (\texttt{c1}–\texttt{c4}, \texttt{f1}–\texttt{f4} and compound clefs such as \texttt{f1/c3} for sources with letter clefs on several lines), staff positions (\texttt{l} for a line, \texttt{z} for a space, with an index counted from the bottom line; \texttt{z-1} and \texttt{l-1} below the staff), accidentals (\texttt{b\_\allowbreak{}} and \texttt{na\_\allowbreak{}} with a position), a separator \texttt{,} between syllable groups and a division mark \texttt{L}. Liquescent shapes and custodes are not encoded. The CTC head uses 59 output tokens, the decoder 68. Figure 1 shows a staff with its target.

\begin{figure}[tbp]
\centering
\includegraphics[width=\textwidth]{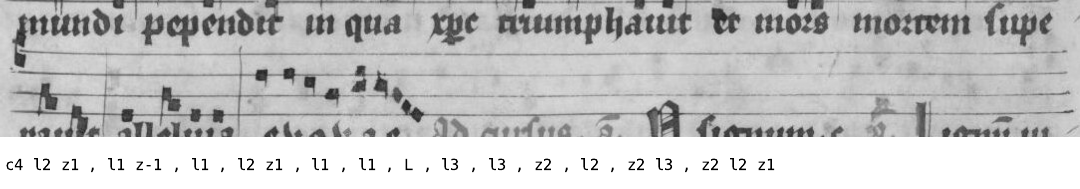}
\caption{A staff of the fixed test set (CH-E 611, fol. 223v, staff 11) with its target. The C clef on the fourth line (\texttt{c4}) is followed by notes on lines (\texttt{l}) and spaces (\texttt{z}), separators between syllable groups (\texttt{,}) and a division mark (\texttt{L}).}
\end{figure}

\subsection{Corpora}
\begin{table}[tbp]
\centering
\small\setlength{\tabcolsep}{4pt}
\caption{Training and evaluation data.}
\begin{tabularx}{\textwidth}{>{\raggedright\arraybackslash\hspace{0pt}}X>{\raggedright\arraybackslash\hspace{0pt}}X>{\raggedright\arraybackslash\hspace{0pt}}X>{\raggedright\arraybackslash\hspace{0pt}}X}
\toprule
Corpus & Sources & Size & Use \\
\midrule
Cantus-aligned ground truth & 10 manuscripts in 11 exports (D-KA Aug. LX main part and addendum), about 1100 to 15th century & export: 26,378 train, 3,109 validation, 3,084 test staves, 175 excluded; folio-disjoint. Training index of Section 6.1: 26,012 unique staves (export plus harvested lines), 32,123 entries after oversampling & all models; folds \\
Transkribus ground truth, reviewed & 22 chant books, mostly German Gothic notation (without the five held-out manuscripts of Section 6.4) & 3,628 unique staves, 10,884 entries after oversampling & CRNN of Section 6.4 (index: 43,877 training entries, 30,445 unique staves in all) \\
AMNLT, staff level & Einsiedeln, Salzinnes, Solesmes & 1,816 / 2,939 / 847 staves, published splits; after conversion to our encoding 1,739 / 2,911 / 838 (train 1,215 / 2,034 / 586, validation 342 / 583 / 166, all 182 / 294 / 86 test staves kept) & AMNLT benchmark \\
Melody corpus from the Cantus Database & 36,567 melody records & 193,479 line sequences (183,881 train, 9,598 validation) & melody model, decoder initialisation \\
\bottomrule
\end{tabularx}
\end{table}

The Cantus-aligned ground truth was derived in earlier work by aligning Volpiano transcriptions of the Cantus Database with staff images of ten manuscripts in eleven exports \cite{klugseder2026}; its split assigns whole folios to training, validation or test. The earlier report counts 3,086 test staves; two staves were removed in a later correction of the export, and the test folios have not changed since. Lines harvested later by a model are used only in training and are excluded when they lie on a validation or test folio. The Transkribus ground truth was transcribed by hand in an earlier project and reviewed for this work; the review procedure will be described separately. The melody corpus contains the chants of the Cantus Database with a full Volpiano melody, excluding the Old Roman source I-Rvat San Pietro B.79 and the 5,749 chants on the validation and test folios of the Cantus-aligned ground truth (with overflow from the preceding folio); they are cut into lines and converted into staff-position tokens, and line sequences with tokens outside the vocabulary are dropped, leaving 36,567 chant records with 193,479 line sequences.

The reference of the Cantus-aligned ground truth is not error-free. In the earlier report, a chant scholar judged samples of the exported staves of each source on the image (Table 2): errors were found in 0 to 4 staves per sample of 34 to 210, with 95 \% Wilson upper bounds between 2.6 and 14.5 \%; for A-KN 1010, 16 of 101 verses were assigned to the wrong staff. A comparison of CH-E 611 with the AMNLT transcription of the same manuscript found that 357 of 1,519 matched staves differ between the two, and that 13 of 39 judged differences contain an error of our reference; assuming agreeing staves to be correct, about 7.8 \% of the staves of CH-E 611 contain a reference error \cite{klugseder2026}. The samples are conditional and the error definitions differ between sources, and a rate of erroneous staves is not a lower bound of the token error. Part of the residual error on these sources may therefore be reference error.

\begin{table}[tbp]
\centering
\small\setlength{\tabcolsep}{4pt}
\caption{Spot checks of the reference by source (staves with an error / staves judged on the image, Wilson upper bound), from \cite{klugseder2026}.}
\begin{tabularx}{\textwidth}{>{\raggedright\arraybackslash\hspace{0pt}}X>{\raggedright\arraybackslash\hspace{0pt}}X>{\raggedright\arraybackslash\hspace{0pt}}X}
\toprule
Source & Errors / judged & Upper bound \\
\midrule
A-KN 1010 & 16 of 101 verses misassigned & – \\
A-KN 1012 & 4 / 206 & 4.9 \% \\
A-KN 1018 & 1 / 202 & 2.8 \% \\
A-Wn 1799 & 0 in 34 certain judgments, 1 further unclear case & 14.5 \% (conservative, for 1 / 35) \\
CH-E 611 & 2 / 99 & 7.1 \% \\
CH-Fco 2 & 0 / 65 & 5.6 \% \\
CH-P 18 & 2 / 200 & 3.6 \% \\
D-KA Aug. LX & 1 / 81 & 6.7 \% \\
D-KA Aug. LX addendum & 0 / 70 (pitches) & 5.2 \% \\
D-KNd 1161 & 3 / 85 & 9.9 \% \\
F-Pn lat. 12044 & 1 / 210 & 2.6 \% \\
\bottomrule
\end{tabularx}
\end{table}

\subsection{Test sets and leakage control}
\begin{table}[tbp]
\centering
\small\setlength{\tabcolsep}{4pt}
\caption{Test sets.}
\begin{tabularx}{\textwidth}{>{\raggedright\arraybackslash\hspace{0pt}}X>{\raggedright\arraybackslash\hspace{0pt}}X>{\raggedright\arraybackslash\hspace{0pt}}X>{\raggedright\arraybackslash\hspace{0pt}}X}
\toprule
Test set & Size & Seen in training & Used in \\
\midrule
Fixed test set & 3,084 staves, 11 exports, 330 source–folio pairs & other folios of the same sources & Sections 6.1, 6.4 \\
Leave-one-source-out & whole source, 1,175 to 7,413 staves & no images and no melody records of the source; other manuscripts' versions of the same chants remain (see below) & Section 6.2 \\
Five held-out manuscripts & 200 staves, 40 each & not by the CRNN of Section 6.4 nor by the vision-language model & Section 6.4 \\
AMNLT test & 182 / 294 / 86 staves & see below & Section 6.3 \\
\bottomrule
\end{tabularx}
\end{table}

We checked the data for leakage before measuring. The index of the Cantus-aligned ground truth is folio-disjoint (3,512 source–folio pairs after merging harvest suffixes, none shared between splits). The fine-tuning corpus of the vision-language model inherits the same split; one of its multi-staff training examples lies on a test folio (CH-P 18, p. 55), so the two staves of that folio are removed from the shared test set. 168 of the 182 AMNLT Einsiedeln test staves lie on training folios of CH-E 611, which is the same manuscript; every model trained on CH-E 611 is affected, so the CRNN is evaluated on Einsiedeln with the model of the fold without CH-E 611, and the corresponding value of the vision-language model is marked. Salzinnes is itself a Cantus Database source (2,052 melodies), so its melody model is retrained without it. The AMNLT splits are not page-disjoint; this applies to published results and ours alike. In the folds, the held-out manuscript contributes no images and no melody records to training, while other manuscripts' versions of the same chants remain available, so the folds test unseen manuscripts, not unseen chants. One aggregate exception remains: the clef frequencies and line lengths that shape the synthetic segments of the melody corpus were computed from all staves of the Cantus-aligned ground truth, including the held-out source. They are corpus-level statistics of a few numbers and contain no melody, but the folds are not strictly free of the held-out source in this respect.

\section{Model}
\subsection{Architecture}
Figure 2 gives an overview. The encoder takes a grey-scale staff image of height 128. Four 3$\times$3 convolutions with 32, 64, 128 and 256 channels and pooling (2,2), (2,2), (2,1), (2,1) produce 2,048 features per column, which a two-layer bidirectional LSTM with 256 units per direction turns into a sequence. A linear CTC head maps each step to 59 tokens and a blank. For the decoder, the encoder output is projected to 256 dimensions and given learned positions; a six-layer Transformer decoder (model width 256, four heads, feed-forward 1,024, pre-norm, output layer tied to the token embedding) writes at most 230 tokens. The model has 14.30 million parameters (encoder 6.69 M, decoder side 7.58 M). The melody model is a causal Transformer of the same shape with a context of 224 tokens and 4.81 M parameters, trained to predict the next token of the melody corpus.

\begin{figure}[tbp]
\centering
\includegraphics[width=\textwidth]{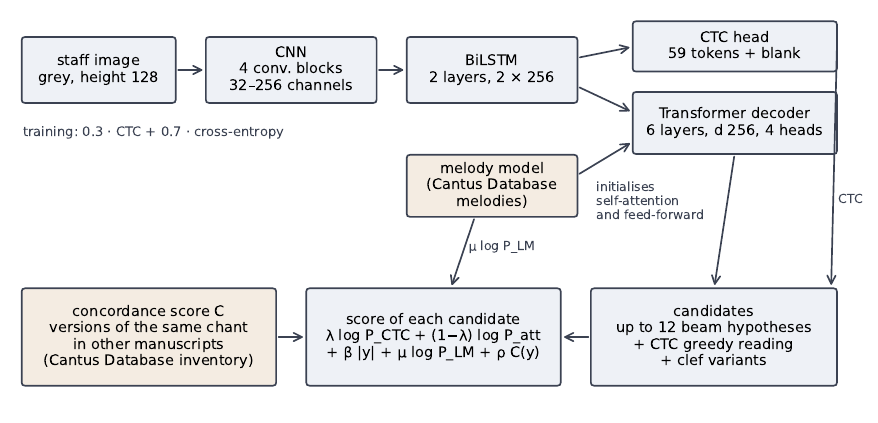}
\caption{Model and decoding. Upper row: the recogniser; the melody model initialises the decoder. Lower row: candidates and their score at reading time; the melody model and the concordance score are the repertoire priors.}
\end{figure}

\subsection{Training}
The loss is 0.3 $\cdot$ CTC + 0.7 $\cdot$ cross-entropy with label smoothing 0.1; AdamW with weight decay 0.01, a one-cycle schedule with 5 \% warm-up, gradient clipping at 5.0, batch size 16, bfloat16 mixed precision, and image augmentation by perspective, affine and elastic distortion, morphology, contrast and brightness. The checkpoint with the lowest attention validation error on a fixed subset of 800 validation staves is kept; for the CTC-only model, the lowest CTC validation error. When the encoder is taken over from the earlier CRNN, its LSTM is used with batch-first input, which the original code had set inconsistently. With a pre-trained decoder, token and position embeddings, self-attention, feed-forward and the corresponding normalisation layers are copied from the melody model; cross-attention is new. A neutral decoder starts from random weights.

\begin{table}[tbp]
\centering
\small\setlength{\tabcolsep}{4pt}
\caption{Training recipes.}
\begin{tabularx}{\textwidth}{>{\raggedright\arraybackslash\hspace{0pt}}X>{\raggedright\arraybackslash\hspace{0pt}}Xl>{\raggedright\arraybackslash\hspace{0pt}}X>{\raggedright\arraybackslash\hspace{0pt}}X}
\toprule
Recipe & Start & Epochs & Learning rates (encoder / pre-trained / rest) & Data \\
\midrule
Base (Section 6.1) & encoder and CTC head from an earlier CRNN trained on the same split & 25 & 2e-4 / 4e-4 / 4e-4 & Cantus-aligned \\
Fold (Section 6.2) & random encoder & 40 & 1e-3 / 4e-4 / 4e-4 & Cantus-aligned without the held-out source \\
General (Section 6.4) & base model of the project (\texttt{hybrid\_\allowbreak{}hybrid\_\allowbreak{}lang}, trained on the Cantus-aligned ground truth) & 10 & 1e-4 for all & Cantus-aligned + 22 Transkribus books \\
AMNLT, pre-trained (Section 6.3) & base or fold model & 30 & 1e-4 for all & AMNLT training split \\
AMNLT only (Section 6.3) & random & 100 & 1e-3 / – / 4e-4, neutral decoder & AMNLT training split \\
\bottomrule
\end{tabularx}
\end{table}

\subsection{Decoding and repertoire priors}
Candidates for a staff are up to twelve completed hypotheses of a beam search scored by the attention decoder alone, and the greedy CTC reading; for the three best hypotheses and the CTC reading, the initial clef is in addition replaced by every clef of the vocabulary. CTC, melody model and concordances do not guide the search; they rescore the completed candidates (N-best rescoring). Each candidate y receives
\begin{equation*}
s(y) = \lambda \log P_{\mathrm{CTC}}(y \mid x) + (1-\lambda) \log P_{\mathrm{att}}(y \mid x) + \beta\,|y| + \mu \log P_{\mathrm{LM}}(y) + \rho\, C(y),
\end{equation*}
where $P_{\mathrm{LM}}$ is the unchanged melody model, applied to the first 222 tokens of a candidate (its context is 224 including start and end), and C the concordance score. For C, the chants on the folio of the staff (and the last chant of the previous folio) are taken from the Cantus Database inventory of the source, and the distinct melodies of their versions in other manuscripts (with at least four notes, excluding San Pietro B.79) are collected; the source itself is excluded. The candidate's pitches, computed from its own clef with repeated notes merged, are compared with each version by the best partial-string similarity (RapidFuzz \texttt{partial\_\allowbreak{}ratio}), and C(y) is the best similarity divided by 100 and multiplied by the number of note tokens of the candidate. The concordance score is gated: a baseline reading is chosen with 0.3 $\cdot$ log $P_{\mathrm{CTC}}$ + 0.7 $\cdot$ log $P_{\mathrm{att}}$ alone, and for every candidate the offset d of its first pitch from the first pitch of the baseline is computed. Each distinct version votes for the shift between $-$7 and +7 diatonic steps under which the baseline best matches it (ties favour the smallest absolute shift and then the negative shift; ties in vote counts follow the same rule), and m is the shift with most votes. A candidate with d $\neq$ 0 receives C only if d = m (candidates without an initial clef yield no pitches and receive C = 0). The gate is thus defined by the first pitch, not by the clef; it is triggered mostly by clef changes, but also by a changed first note. The weights were chosen on the validation set of the fixed split ($\lambda$ = 0.3, $\beta$ = 0.5, $\mu$ = 0.05, $\rho$ = 0.5) and are used unchanged in every experiment. ``Joint'' in the tables means rescoring this full candidate set (beam 12 plus greedy CTC and clef variants) with $\mu$ = $\rho$ = 0. Sources without a folio inventory in the Cantus Database (the Transkribus books, the five held-out manuscripts, AMNLT) are read without concordances.

\section{Experimental protocol}
\textbf{Metrics.} The token error rate (symbol error rate, SER) is the Levenshtein distance between token sequences divided by the length of the reference, micro-averaged over staves ($\Sigma$ distance / $\Sigma$ reference length); the same holds for the other rates. Melody SER compares pitches obtained from the clef that was read, without separators, bars and accidentals and with repeated notes merged; staves without an initial clef in the reference are left out. A wrong clef transposes every pitch of the staff, which an edit distance counts as many substitutions. For comparisons with the vision-language model and on AMNLT we use the pitch-position SER of the LightOn article, the Levenshtein distance between the sequences of staff positions, so that both models are measured identically \cite{klugseder2026lighton}. On AMNLT, pitch-position SER is evaluated against the pitch sequence of the original GABC target for all 182 / 294 / 86 test staves, including 12 Einsiedeln and 5 Salzinnes test staves whose accidentals fall outside our token vocabulary and which carry placeholder labels in the token index. Initial-clef accuracy is defined in two ways. In Sections 6.1 and 6.2 it compares the effective clef, the pitch the clef gives to the bottom line, so that \texttt{c4} and \texttt{f2} count as the same. In Sections 6.3 and 6.4, as in the LightOn article, the clef that was read must be one of the clefs written in the reference.

\textbf{Seeds and uncertainty.} Sections 6.1 and 6.4 use three seeds; in Section 6.1 all runs start their encoder from the same earlier CRNN, and in Section 6.4 from the same base model, so the seeds vary the new parts and the batch order and understate the variance of a full retraining. Section 6.2 uses three seeds for the three folds where the priors had an effect in seed 0, a choice made after seeing that result, and one seed for the other three. Differences between readings are tested by a paired bootstrap with 2,000 resamples and a percentile 95 \% interval. The resampling unit is the identified source–folio pair (fixed test set and folds; 330 pairs on the fixed test set) or the identified source–page pair (Section 6.4), because staves of one page are not independent. The AMNLT test staves carry no page identifier, so Section 6.3 resamples staves and ignores possible within-page dependence; the coverage of these intervals is uncertain and they may understate uncertainty. Three comparisons were named in the measurement plan before the runs (hybrid with joint rescoring against CTC only, pre-trained against neutral decoder, CRNN against the vision-language model); we report all intervals as pointwise and descriptive, without correction for multiple comparisons, conditional on the evaluated sources and trained checkpoints. Consistent signs across seeds show stability under these checkpoints, not independent replications across manuscripts. ``No measurable difference'' means that the interval includes zero; it does not establish equivalence.

\textbf{Selection.} Checkpoints and decoding weights were chosen on validation data only. Two later analyses are exploratory and were developed after inspecting test results: the additional seeds of three folds and the clef lock of Section 6.2; they are evaluated on the same test data and marked as such.

\textbf{Image check.} Where the priors change the initial clef, a chant scholar (the author) judged on the image, blind to the origin of the alternatives, which clef stands at the beginning of the staff (Section 6.2).

\textbf{Compute.} Each run used one NVIDIA H100; training took about 4 hours (Section 6.1) and 4.5 to 6.7 hours (Section 6.2) according to the job logs; at batch 16 the logs show the GPU partly idle, which we attribute to the data-loading main process (not profiled). Reading times were measured separately (Section 6.4).

\section{Results}
\subsection{Ablation on the fixed test set}
\begin{table}[tbp]
\centering
\footnotesize\setlength{\tabcolsep}{4pt}
\caption{Token SER (\%) on the fixed test set (3,084 staves), mean $\pm$ sd over three seeds.}
\begin{tabularx}{\textwidth}{>{\raggedright\arraybackslash\hspace{0pt}}Xllllll}
\toprule
Model & CTC greedy & Attention greedy & Joint & + melody model & + concordances & both \\
\midrule
CTC only & 7.10 $\pm$ 0.04 & – & – & – & – & – \\
Hybrid, neutral decoder & 7.20 $\pm$ 0.03 & 6.78 $\pm$ 0.15 & 5.84 $\pm$ 0.01 & 5.82 $\pm$ 0.03 & 5.84 $\pm$ 0.02 & 5.82 $\pm$ 0.02 \\
Hybrid, pre-trained decoder & 7.24 $\pm$ 0.02 & 6.37 $\pm$ 0.13 & 5.65 $\pm$ 0.05 & 5.63 $\pm$ 0.07 & 5.66 $\pm$ 0.04 & 5.64 $\pm$ 0.05 \\
\bottomrule
\end{tabularx}
\end{table}

The largest single step is from the CTC-only model to the hybrid with joint rescoring: 7.10 against 5.84 \% with a neutral decoder (seed means), a package of hybrid training, beam search and rescoring whose parts are not separated here. For this comparison we report seed means only; no paired interval is reported. Initialising the decoder from the melody model lowers the joint SER by 0.25, 0.15 and 0.17 points in the three seed pairs (intervals \mbox{[$-$0.34, $-$0.17]}, \mbox{[$-$0.24, $-$0.06]}, \mbox{[$-$0.27, $-$0.07]}) and the melody SER by 0.54, 0.42 and 0.61 points (intervals \mbox{[$-$0.89, $-$0.23]}, \mbox{[$-$0.85, +0.01]}, \mbox{[$-$1.02, $-$0.21]}). On these known sources the priors make no measurable difference: with both priors the token SER changes by $-$0.00 \mbox{[$-$0.04, +0.05]} and the melody SER by $-$0.00 \mbox{[$-$0.20, +0.19]} points (seed 0). Where they change the initial clef (21 of 3,084 staves), 5 changes turn an incorrect clef into the reference clef, 13 turn a correct clef into an incorrect one, and 3 replace one incorrect clef by another.

\subsection{Unseen manuscripts}
\begin{table}[tbp]
\centering
\footnotesize\setlength{\tabcolsep}{4pt}
\caption{Leave-one-source-out, token SER (\%), seed 0.}
\begin{tabularx}{\textwidth}{>{\raggedright\arraybackslash\hspace{0pt}}Xllllll}
\toprule
Held-out source & Staves & CTC only & Joint & + melody model & + concordances & both \\
\midrule
CH-E 611 & 7,413 & 4.13 & 3.39 & 3.33 & 3.41 & 3.34 \\
A-Wn 1799 & 3,137 & 12.01 & 10.48 & 10.22 & 10.72 & 10.47 \\
A-KN 1012 & 1,175 & 15.96 & 13.09 & 13.07 & 13.04 & 12.97 \\
F-Pn lat. 12044 & 2,660 & 15.60 & 13.16 & 13.15 & 13.05 & 13.07 \\
CH-P 18 & 1,826 & 13.25 & 14.43 & 14.32 & 14.49 & 14.40 \\
D-KA Aug. LX with addendum & 4,246 & 23.47 & 23.31 & 23.21 & 23.51 & 23.41 \\
\bottomrule
\end{tabularx}
\end{table}

Joint decoding helps in four of six folds (by up to 2.9 points) but not on CH-P 18, where it is 1.2 points worse than CTC alone, and hardly on D-KA Aug. LX, which stays at 23 \% in every configuration (Section 6.5). The priors move the token SER by at most 0.26 points. Their effect lies in the melody and the clef (Table 7).

\begin{table}[tbp]
\centering
\footnotesize\setlength{\tabcolsep}{4pt}
\caption{Melody SER (\%) without and with both priors, bootstrap over source–folio pairs.}
\begin{tabularx}{\textwidth}{>{\raggedright\arraybackslash\hspace{0pt}}Xll>{\raggedright\arraybackslash\hspace{0pt}}X>{\raggedright\arraybackslash\hspace{0pt}}X>{\raggedright\arraybackslash\hspace{0pt}}X}
\toprule
Held-out source & Folios & Seed & Melody SER joint $\rightarrow$ both & $\Delta$ [95 \%] & Clef transitions: incorrect$\rightarrow$correct / correct$\rightarrow$incorrect \\
\midrule
A-Wn 1799 & 373 & 0 & 23.64 $\rightarrow$ 21.29 & $-$2.35 \mbox{[$-$3.54, $-$1.18]} & 122 / 52 \\
 &  & 1 & 23.38 $\rightarrow$ 20.96 & $-$2.42 \mbox{[$-$3.80, $-$1.16]} & 150 / 84 \\
 &  & 2 & 23.35 $\rightarrow$ 22.25 & $-$1.10 \mbox{[$-$2.12, $-$0.03]} & 103 / 62 \\
A-KN 1012 & 264 & 0 & 19.06 $\rightarrow$ 17.35 & $-$1.70 \mbox{[$-$2.65, $-$0.86]} & 21 / 1 \\
 &  & 1 & 20.00 $\rightarrow$ 18.11 & $-$1.89 \mbox{[$-$2.80, $-$1.09]} & 21 / 0 \\
 &  & 2 & 18.91 $\rightarrow$ 17.56 & $-$1.35 \mbox{[$-$2.18, $-$0.61]} & 14 / 2 \\
F-Pn lat. 12044 & 301 & 0 & 30.97 $\rightarrow$ 27.30 & $-$3.67 \mbox{[$-$4.53, $-$2.80]} & 117 / 15 \\
 &  & 1 & 29.59 $\rightarrow$ 26.40 & $-$3.19 \mbox{[$-$4.06, $-$2.36]} & 112 / 23 \\
 &  & 2 & 23.41 $\rightarrow$ 21.00 & $-$2.42 \mbox{[$-$3.19, $-$1.64]} & 94 / 21 \\
CH-E 611 & 514 & 0 & 4.90 $\rightarrow$ 4.70 & $-$0.20 \mbox{[$-$0.39, $-$0.02]} & 26 / 16 \\
CH-P 18 & 257 & 0 & 20.22 $\rightarrow$ 20.08 & $-$0.14 \mbox{[$-$1.08, +0.78]} & 31 / 30 \\
D-KA Aug. LX with addendum & 428 & 0 & 33.66 $\rightarrow$ 33.15 & $-$0.52 \mbox{[$-$0.91, $-$0.14]} & 52 / 20 \\
\bottomrule
\end{tabularx}
\end{table}

In the three early manuscripts A-Wn 1799, A-KN 1012 and F-Pn lat. 12044, both priors together lower the melody SER in every seed, by 1.96 $\pm$ 0.74, 1.65 $\pm$ 0.27 and 3.09 $\pm$ 0.63 points (1.10 to 3.67 in single runs), and the concordance score alone accounts for most of it. The gain comes mostly through the initial clef: in these folds, the clefs the priors change are changed from incorrect to correct far more often than the reverse. The seeds also show how unstable a single run is under a change of manuscript: without priors, the melody SER of F-Pn lat. 12044 ranges from 23.4 to 31.0 \% across seeds, and the token SER of A-Wn 1799 from 9.3 to 11.6 \%.

\textbf{Image check.} Both the reference and the concordances come from the Cantus Database, so agreement with the reference could reflect shared editorial conventions rather than the image. We therefore had the changed clefs judged on the image (Section 5): all 21 changes on the fixed test set, all 23 in A-KN 1012 and 30 sampled changes each in F-Pn lat. 12044 and A-Wn 1799.

\begin{table}[tbp]
\centering
\small\setlength{\tabcolsep}{4pt}
\caption{Changed initial clefs judged on the image.}
\begin{tabularx}{\textwidth}{>{\raggedright\arraybackslash\hspace{0pt}}Xl>{\raggedright\arraybackslash\hspace{0pt}}Xll}
\toprule
Test set & Judgeable & Reference correct & Prior corrects & Prior spoils \\
\midrule
Fixed test set (known sources) & 17 of 21 & 15 & 5 & 9 \\
A-KN 1012 (unseen) & 17 of 23 & 16 & 14 & 1 \\
F-Pn lat. 12044 (unseen) & 30 of 30 & 29 & 27 & 1 \\
A-Wn 1799 (unseen) & 30 of 30 & 29 & 17 & 12 \\
\bottomrule
\end{tabularx}
\end{table}

The reference has the correct clef in 89 of the 94 judgeable staves. The image check therefore argues against reference-clef errors as the sole explanation of the gains in these samples; it does not validate the full melody reference, and it covers only staves whose clef was changed. In the examined samples, image judgments corroborate substantial gains for A-KN 1012 and F-Pn lat. 12044 and a smaller net gain for A-Wn 1799 (17 against 12), while on the known sources the priors caused more clef errors than they corrected (9 against 5). A second, exploratory sample concerns the melody model alone: in a transcription of 25 further, model-known manuscripts with the general model and the melody model (no concordances, since these sources have no folio inventory), 1,015 of 86,166 readings had a changed initial clef. Of 60 sampled staves, 23 have no clef written at their beginning, and the melody model had added one; of the 32 with a visible clef, the clef without the melody model was correct in 16, the changed one in 14, neither in 2; 5 were not judgeable (3 with the beginning cut off, 2 crops without a staff). Against the image, the reading without the melody model is thus right in 39 of 55 judgeable staves, mostly because the melody model writes clefs the scribe did not write; where a clef is visible, both readings perform similarly (16 against 14). In an exploratory follow-up, we held the clef of the reading without priors fixed and let the priors choose only among candidates with that clef. This keeps the token SER of the fixed test set at or slightly below that of free priors ($-$0.03, $-$0.02 and $-$0.03 points, intervals excluding zero) and removes the clef changes, but on the unseen manuscripts it gives up the melody gain (+1.5 to +3.5 points). We therefore propose source familiarity as a deployment heuristic: let the priors change the clef only for sources the model has not seen. Its performance on further manuscripts remains to be validated.

\subsection{AMNLT benchmark}
\begin{table}[tbp]
\centering
\footnotesize\setlength{\tabcolsep}{4pt}
\caption{AMNLT test sets, pitch-position SER (\%) with initial-clef accuracy (\%) in parentheses. LightOnOCR P-d is the default fine-tuned configuration of the LightOn article (single staves and multi-staff images, AMNLT, Cantus-aligned and Transkribus gold and agreement-filtered silver data) \cite{klugseder2026lighton}.}
\begin{tabularx}{\textwidth}{llll>{\raggedright\arraybackslash\hspace{0pt}}Xl}
\toprule
Dataset & Staves & LightOnOCR P-d & CRNN, AMNLT only & CRNN, pre-trained (base) & + melody model \\
\midrule
Einsiedeln & 182 & (1.10, 100.0) * & 2.55 (99.5) & 1.40 (99.5) (fold without CH-E 611) & 1.41 \\
Salzinnes & 294 & 1.41 (99.7) & 1.64 (99.7) & 1.44 (98.6) (decoder without Salzinnes melodies) & 1.43 \\
Solesmes & 86 & 1.23 (100.0) & 10.21 (73.3) & 2.43 (93.0) (base, seed 0) & 2.49 \\
\bottomrule
\end{tabularx}
\par\smallskip{\footnotesize * trained on CH-E 611, which overlaps the test staves (Section 3.3); excluded from the comparison.}
\end{table}

Pre-training on our corpora matters most for the small Solesmes set (10.21 $\rightarrow$ 2.43 \%). On Salzinnes we did not detect a difference from the vision-language model ($-$0.03 \mbox{[$-$0.25, +0.22]} points); on Solesmes the CRNN is 1.20 \mbox{[0.66, 1.78]} points behind. These intervals resample staves rather than pages and should be interpreted with that limitation (Section 5). Because the base models differ by dataset, differences between datasets also reflect the base. The published AMNLT music error rates include shape signs that the CRNN does not output and are therefore not directly comparable; the LightOn article relates the vision-language model to them \cite{klugseder2026lighton}.

\subsection{Comparison with a fine-tuned vision-language model}
For the comparison, the CRNN was trained like the project's general model, on the Cantus-aligned ground truth and the reviewed Transkribus books, but without the five held-out manuscripts that the vision-language model has never seen either (Table 4, ``General''). Its training index contains 32,993 non-Transkribus entries (26,817 distinct staves; it includes later additions for A-KN 1012 and correction exclusions) and 10,884 entries for 3,628 distinct staves from 22 Transkribus books, so it is not simply the index of Section 6.1 with the books appended. The comparison and its recipe were fixed before the run. Both models read their own crop of the same staff; reference and metric are shared, the training data are not (Table 1 and the LightOn article).

\begin{table}[tbp]
\centering
\footnotesize\setlength{\tabcolsep}{4pt}
\caption{CRNN against LightOnOCR-3-1B (P-d), pitch-position SER (\%).}
\begin{tabularx}{\textwidth}{>{\raggedright\arraybackslash\hspace{0pt}}Xll>{\raggedright\arraybackslash\hspace{0pt}}X>{\raggedright\arraybackslash\hspace{0pt}}X>{\raggedright\arraybackslash\hspace{0pt}}X}
\toprule
Test set & Staves & LightOn & CRNN, seeds 0 / 1 / 2 & $\Delta$ LightOn $-$ CRNN (seed 0) [95 \%] & Initial clef LightOn / CRNN \\
\midrule
Known: fixed test set $\cap$ LightOn test split & 1,041 & 4.47 & 4.50 / 4.50 / 4.56 & $-$0.02 \mbox{[$-$0.47, +0.50]} & 96.7 / 96.6–96.7 \\
Unseen: five held-out manuscripts & 200 & 1.31 & 1.80 / 1.84 / 1.82 & $-$0.49 \mbox{[$-$1.03, +0.06]} & 91.0 / 88.5–90.0 \\
\bottomrule
\end{tabularx}
\end{table}

On the shared known-source subset we did not detect a difference; the 95 \% interval permits an advantage of about half a point for either system. On the five unseen manuscripts the vision-language model is 0.5 points ahead, with an interval that just includes zero; the interval is conditional on these five manuscripts, of which the pages are the resampled unit, and allows no statement about unseen manuscripts in general; per manuscript the CRNN is ahead on three (A-Geras 6 0.16 against 0.49 \%, A-KN 65 1.90 against 2.28 \%, A-Wn 15484 2.00 against 2.43 \%) and behind on two (A-Gu 17 3.20 against 1.28 \%, A-VOR 254 1.83 against 0.76 \%). The known set consists of the staves that passed the crop filter of the vision-language model, a cleaner subset of the fixed test set; its values are not comparable with the token SER of Table 5. Both models are read in the standard setting of their project (the CRNN with joint decoding and weights from validation, the vision-language model greedily with a checkpoint selected on its validation data); part of the silver training data of the vision-language model was selected by agreement with a CRNN reading, so the two are not trained independently.

Adding reviewed Gothic training data, together with the associated fine-tuning, reduced the CRNN's error on the unseen Gothic manuscripts substantially: trained on the Cantus-aligned ground truth alone, the same architecture reads the five held-out manuscripts with 8.2 to 10.3 \% pitch-position SER, against 1.8 \% with the Transkribus books. This experiment does not isolate an effect of model size, and the corresponding ablation for the vision-language model was not run. The CRNN has 14.3 million parameters, the vision-language model about one billion. Reading times were measured on one H100, in a single process with batch size 1, on the 200 unseen staves. The CRNN timings exclude image loading; the LightOn timing includes image loading, preprocessing, generation, decoding and output parsing, so the two timings do not use identical measurement boundaries. The CRNN reads a staff in 6.7 ms with greedy CTC; a separate beam-6 timing configuration, including melody-model rescoring but excluding the clef variants of the full pipeline, averaged 194 ms per staff (median 188 ms); its accuracy was not evaluated in Table 10. The vision-language model, read greedily as in Table 10, took 1.75 s per staff (median 1.72 s), producing on average 102 tokens including the lyrics, 17.2 ms per token.

\subsection{Error analysis}
\begin{table}[tbp]
\centering
\scriptsize\setlength{\tabcolsep}{3pt}
\caption{Error types of the joint reading (pre-trained decoder, no priors), as share of all token errors (\%), from one optimal Levenshtein alignment per staff. Last column: share of the missing notes, not of all errors.}
\begin{tabularx}{\textwidth}{>{\raggedright\arraybackslash}p{2.2cm}>{\raggedright\arraybackslash\hspace{0pt}}X>{\raggedright\arraybackslash\hspace{0pt}}X>{\raggedright\arraybackslash\hspace{0pt}}X>{\raggedright\arraybackslash\hspace{0pt}}X>{\raggedright\arraybackslash\hspace{0pt}}X>{\raggedright\arraybackslash\hspace{0pt}}X>{\raggedright\arraybackslash\hspace{0pt}}X>{\raggedright\arraybackslash\hspace{0pt}}X>{\raggedright\arraybackslash\hspace{0pt}}X>{\raggedright\arraybackslash\hspace{0pt}}X}
\toprule
Test set & Token SER & Note missing & Note extra & Position $\pm$1 & Position $\pm$2 or more & Clef & Separator & Division mark & Accidental & Missing notes that repeat the previous note \\
\midrule
Fixed test set (seed 0) & 5.59 & 18.0 & 15.3 & 8.1 & 17.1 & 1.9 & 32.7 & 4.5 & 2.3 & 26.6 \\
CH-E 611 & 3.39 & 17.6 & 11.3 & 15.0 & 3.2 & 3.7 & 35.5 & 9.4 & 4.4 & 14.9 \\
A-Wn 1799 & 10.48 & 9.9 & 41.2 & 4.6 & 14.3 & 7.6 & 18.6 & 2.8 & 1.1 & 24.5 \\
A-KN 1012 & 13.09 & 30.4 & 12.4 & 17.4 & 5.5 & 1.5 & 26.6 & 5.0 & 1.2 & 27.9 \\
F-Pn lat. 12044 & 13.16 & 29.5 & 8.9 & 3.9 & 13.6 & 8.0 & 27.7 & 5.8 & 2.7 & 21.6 \\
CH-P 18 & 14.43 & 22.9 & 19.1 & 6.2 & 11.2 & 6.4 & 28.4 & 4.3 & 1.5 & 23.4 \\
D-KA Aug. LX with addendum & 23.31 & 13.9 & 34.9 & 4.9 & 10.0 & 4.5 & 28.5 & 2.9 & 0.5 & 24.3 \\
\bottomrule
\end{tabularx}
\end{table}

On the fixed test set, a third of all token errors concern separators between syllable groups, which depend on how the reference divides the melody into syllables rather than on the image alone. Missing and extra notes together account for 33.3 \% of the token errors, against 25.2 \% for wrong positions (two thirds of them by two steps or more), and 26.6 \% of the missing reference notes repeat the preceding staff position. These shares describe edit operations of one alignment, not observed palaeographic causes. Per source, the token SER ranges from 1.5 \% (CH-Fco 2) and 1.9 \% (CH-E 611) to 12.6 \% (D-KA Aug. LX) and 16.6 \% (A-KN 1010). The two folds where joint decoding did not help, CH-P 18 and D-KA Aug. LX, differ in kind. In D-KA Aug. LX extra notes dominate (35 \% of the errors), as they also do in A-Wn 1799 (41 \%); the addendum of D-KA Aug. LX, a later section on five-line staves, reaches 37.9 \% against 22.5 \% for the main part, so the fold is burdened by a staff layout that occurs nowhere else in the training data. CH-P 18 shows no single dominant type. Whether part of the extra notes reflects notes omitted in the Cantus-derived reference was not checked.

\section{Discussion}
The hybrid with joint rescoring has the largest observed mean improvement in our ablation; it lets the attention decoder use what it has learned about the order of notes while the CTC score keeps hypotheses close to the image. Melody pre-training of the decoder helps as an initialisation, consistently but modestly, while the same melody model adds almost nothing as a separate score on known sources. A plausible reading is that a decoder trained on images of a source already contains what the melody model could contribute; this was not tested further.

Repertoire knowledge in our setting is mainly knowledge about clefs and transposition, not about individual notes. This is consistent with Hartelt and Puppe, whose n-gram model of pitches did not improve note recognition while background knowledge reduced clef errors by about a quarter \cite[p.~15]{hartelt2022}: at the level of notes the priors change little, but where the model is unsure about the clef of an unfamiliar source, the versions of the same chant in other manuscripts tell it which reading leads to the melody most of them agree on. In the examined changed-clef samples from known sources, the priors more often harmed than helped; in the production sample, a frequent failure was inserting an initial clef where none was written. These observations motivate the deployment heuristic of Section 6.2, not a validated general rule.

In the comparison with the vision-language model, we did not detect a difference on known manuscripts, with an interval that permits half a point either way. On unseen manuscripts the larger model is somewhat ahead, mainly on two of five manuscripts; with 200 staves and five manuscripts, this does not establish a general advantage. The vision-language model also transcribes lyrics and their alignment, which is outside this article.

\section{Limitations}
The reference of the fixed test set and the folds comes from the Cantus Database, as do both priors; the image check of 104 staves shows the reference initial clef to be correct in 89 of 94 judgeable cases, but it covers only staves with changed clefs. Three of the six folds have one seed; the seeds of Section 6.1 share the encoder initialisation. The unseen set of Section 6.4 has 200 staves from five manuscripts. The folds are limited to the Cantus-aligned sources, which differ in how their reference was derived. The encoding omits liquescence and custodes. The AMNLT splits are not page-disjoint, and the vision-language model's Einsiedeln value is affected by overlap. The CRNN and the vision-language model share part of their silver data. The image check was made by one person, covers only changed clefs, and the clef lock and additional seeds were developed after inspecting test results. The clef and length statistics of the melody corpus include all staves of the Cantus-aligned ground truth (Section 3.3). Image rights limit the release of data.

\section{Conclusion}
For staff-level chant recognition, a hybrid CTC/attention recogniser with joint rescoring showed the largest mean reduction in token SER in our ablation, and initialising the decoder from a model of chant melodies adds a small but consistent gain. Repertoire knowledge at reading time is no general improvement: in the examined samples it corrects clefs of manuscripts the model has not seen and spoils them in manuscripts it knows, which suggests restricting clef changes to unseen sources. With reviewed training data in the target notation, a recogniser of 14 million parameters showed no detectable difference from a fine-tuned vision-language model of about one billion parameters on known manuscripts and trailed it by about half a point on five unseen ones; without such data, its error on those manuscripts is several times higher.

\section*{Availability and disclosure}
The scripts of this study (training, reading with the repertoire priors, evaluation, bootstrap and image checks), the split indices without images and the weights of the seed-0 models of Tables 5 and 10 and of the melody model are available at \url{https://github.com/rklugsederoeaw/omr_cnn_transformer} (folder \texttt{paper\_\allowbreak{}2026}), the code under the Apache License 2.0, weights and indices under CC BY-NC-SA 4.0 in line with the licence of the Cantus Database and the image rights of the training sources. The data loader and the original CRNN, on which the encoder builds, belong to the Austrian Academy of Sciences and were developed on its behalf by the Pattern Recognition and Artificial Intelligence Group (PRAIG) of the University of Alicante (\url{https://github.com/ACDHxUA/CLI}). Staff images and ground truth are published only as far as the rights of the holding libraries permit.

Generative AI tools were used substantially: an AI coding assistant (Claude in Claude Code, Anthropic) wrote and ran the code, the training and evaluation runs and a first draft of this text under the author's direction, and drafted the image-check pages. The author decided all questions of notation, evaluation and interpretation, and made the image checks. Two further AI systems (OpenAI Codex and Google Gemini) reviewed the draft independently and recomputed its numbers from the stored readings; the deviations they found were corrected. The author takes full responsibility for the content.

\end{document}